\documentclass[conference]{IEEEtran}
\IEEEoverridecommandlockouts
\usepackage{cite}
\usepackage{amsmath,amssymb,amsfonts}
\usepackage{algorithmic}
\usepackage{graphicx}
\usepackage{cuted}
\usepackage{textcomp}
\usepackage{xcolor}
\usepackage{caption}
\usepackage{subcaption}

\usepackage[hidelinks]{hyperref}
\hypersetup{breaklinks=true}

\def\BibTeX{{\rm B\kern-.05em{\sc i\kern-.025em b}\kern-.08em
    T\kern-.1667em\lower.7ex\hbox{E}\kern-.125emX}}
\begin{document}

\title{Uncovering the Mechanism of Hepatotoxiciy of PFAS Targeting L-FABP Using GCN and Computational Modeling\\
}

\author{\IEEEauthorblockN{1\textsuperscript{st} Lucas Jividen}
\IEEEauthorblockA{\textit{Department of Computer Science} \\
\textit{University of Cincinnati}\\
Cincinnati, USA \\
jividelh@mail.uc.edu}
\and
\IEEEauthorblockN{2\textsuperscript{nd} Tibo Duran}
\IEEEauthorblockA{\textit{Department of Pharmaceutical Sciences} \\
\textit{University of Connecticut}\\
Storrs, USA \\
tibo.duran@uconn.edu}
\and
\IEEEauthorblockN{3\textsuperscript{rd} Xi-Zhi Niu}
\IEEEauthorblockA{\textit{Department of Environmental Engineering} \\
\textit{University of Cincinnati}\\
Cincinnati, USA \\
niuxh@ucmail.uc.edu}
\and
\IEEEauthorblockN{4\textsuperscript{rd} Jun Bai}
\IEEEauthorblockA{\textit{Department of Computer Science} \\
\textit{University of Cincinnati}\\
Cincinnati, USA \\
baiju@ucmail.uc.edu}
}

\maketitle

\begin{abstract}
Per- and polyfluoroalkyl substances (PFAS) are persistent environmental pollutants with known toxicity and bioaccumulation issues. Their widespread industrial use and resistance to degradation have led to global environmental contamination and significant health concerns. While a minority of PFAS have been extensively studied, the toxicity of many PFAS remains poorly understood due to limited direct toxicological data. This study advances the predictive modeling of PFAS toxicity by combining semi-supervised graph convolutional networks (GCNs) with molecular descriptors and fingerprints. We propose a novel approach to enhance the prediction of PFAS binding affinities by isolating molecular fingerprints to construct graphs where then descriptors are set as the node features. This approach specifically captures the structural, physicochemical, and topological features of PFAS without overfitting due to an abundance of features. Unsupervised clustering then identifies representative compounds for detailed binding studies. Our results provide a more accurate ability to estimate PFAS hepatotoxicity to provide guidance in chemical discovery of new PFAS and the development of new safety regulations.
\end{abstract}


\section{Introduction}

Per and polyfluoroalkyl substances, denoted as PFAS, are a class of chemicals that have gained global attention due to their potential toxicity and persistence in the environment since 1950's. 
PFAS have widespread industrial and commercial uses in food packaging materials, cookware, and firefighting foams \cite{b1,b2}.  Their practicality in industry is due to the physical and chemical stability of these compounds. These compounds typically consist of a non-fluorinated 'head' that is polar and hydrophilic, and a carbon-fluorine 'tail' that is hydrophobic and lipophobic, forming one of the strongest bonds in organic chemistry \cite{b3,b4}.  
The same innate resistance to degradation that made them advantageous in the industrial world, has created hardships in the removal of PFAS from the environment. Thus, PFAS have been documented in various sources around the globe \cite{b5}-\cite{b7}, and human exposure to these compounds is evident in studies such as the National Health and Nutrition Examination Survey (NHANES). The study recorded that greater than 98\% of the US population had detectable concentrations of PFAS based on representative samples \cite{b5}.  
While broader society has become more aware of the adverse health effects of these chemicals, the degree of toxicity of individual PFAS compounds remains unclear, except for some heavily studied legacy PFAS, such as perfluorooctanoic acid (PFOA) and perfluorooctane sulfonate (PFOS ) \cite{b12}. Due to the small concentrations of PFAS in the biological samples, it is not feasible to perform controlled studies to get the direct toxicological effects of specific PFAS. However, epidemiological studies have exposed the relation of PFAS to a plethora of health effects, targeting biological pathways from the liver to lipids \cite{b12}. A significant target for PFAS presents itself as the liver fatty acid binding protein (LFABP). Intrinsically fatty acids have a high binding affinity to LFABP, and PFAS share close structural similarities to such natural fatty acids causing concern for bioaccumulation in these proteins \cite{b13,b14}. This potential binding site has thus been used to study the potential toxicity of PFAS towards its effect on human health \cite{b13}-\cite{b16}, as the tighter the bonding is to a biological target, generally the stronger the toxicity \cite{b17}.

Recently, quantitative structure-activity relationship (QSAR) machine learning models have gained popularity in predicting the potential toxicity of PFAS \cite{b13,b18}. One study, incorporated animal exposure experiments with binding affinities calculated through molecular dynamics (MD) simulations to construct a QSAR model to discover replacements to PFOS with lower binding capability and liver accumulation \cite{b13}. 
Another approach to obtaining the necessary data for untested PFAS was utilized in \cite{b18} which constructed a dataset from available chemical bioactivity data sets that contained carbon fluorine (CF) moieties. The CF bonds are universal with PFAS thus providing valuable structural information. 
Machine learning has also been utilized 
to discover potential PFAS replacements that have reduced toxicity potential \cite{b19}. This study by Lai trained a model on existing PFAS to generate $260,000$ novel compounds as potential replacements for existing PFAS while favoring industrial suitability and lower toxicity. Molecular descriptors with known relationships to the former characteristics were used as indicators to screen the generated molecules before MD simulations. 
Another study also used chemical descriptors to represent PFAS and use as a set of features to predict the binding fraction of PFAS to human serum albumin \cite{b20}. 

The use of chemical descriptors and fingerprints has been a useful tool for translating molecules into a set of features by representing their structural features, physico-chemical, and topological traits. However, the primary issue with predicting the toxicity of PFAS compounds stems from the limited toxicological data for these compounds. To address the problem, we proposed a deep learning and physical modeling hybrid method to predict PFAS toxicity from binding affinity data and uncover binding mechanisms. 
 To our best knowledge, there are no methods that apply PFAS binding affinities data, calculated through MD simulations, with graph convolutional networks utilizing both molecular descriptors and fingerprints. 
In our study, we propose to use a semi-supervised graph convolutional network (GCN) \cite{b21} to capture the structural features, physico-chemical, and topological properties of PFAS through chemical fingerprints and descriptors. We then use an unsupervised clustering method, Ward, to group the global database for PFAS. Select representative PFAS were then chosen from the clusters to analyze their specific binding interactions to LFABP through MD simulations. The code used for this research is available at: \url{https://github.com/LabJunBMI/GCN-PFAS-Docking-Score-Prediction}.

To address the challenge above, we proposed:
\begin{itemize}
    \item employing a GCN model to predict PFAS binding affinity to LFABP with greater accuracy.
    \item utilizing both chemical fingerprints and descriptors to represent the compounds.
    \item using MD simulations to uncover the binding mechanisms of distinct PFAS identified through an unsupervised clustering method.
\end{itemize}

The paper is organized as follows: Section \ref{sec:method} presents the main methods. Section \ref{sec:experiment} summarizes the dataset, data preprocessing, and model/simulation parameters. In Section \ref{sec:result}, results are analyzed and discussed to show performance of the proposed model and insights provided from model predictions and MD simulations. introduces the real datasets and data preprocessing. Section \ref{sec:conclusion} concludes the paper.

\section{Methods}\label{sec:method}

\begin{figure}[htbp]
    \centering
    \includegraphics[width=0.9\columnwidth]{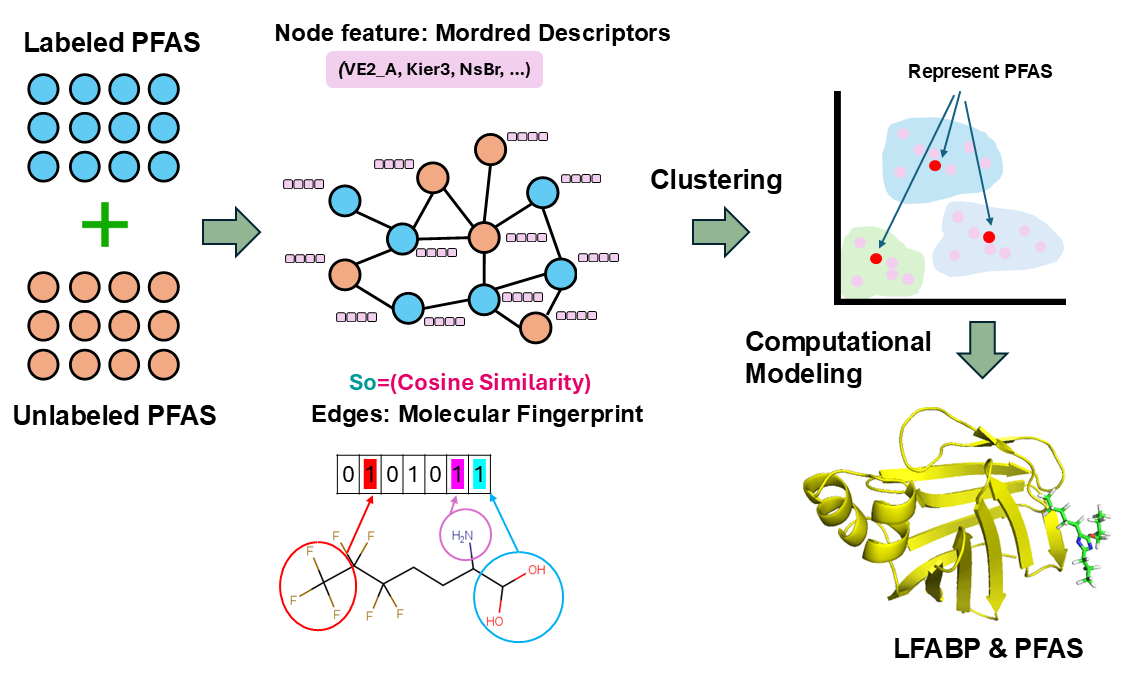}
    \caption{Method Flow Chart}
    \label{fig:method}
    \vspace{-0.1in}
\end{figure}

\subsection{Graph Convolutional Network}

In this section, we present our proposed semi-supervised GCN, which consists of two techniques: graph representation and semi-supervised graph message passing algorithm. The workflow of the proposed method is shown in Figure \ref{fig:method}.

\subsubsection{Definition}

We proposed to employ a GCN model to learn the binding affinity of PFAS. We used an undirected graph to represent existing PFAS and its binding affinity towards LFABP. The graph is defined as follows: \( G = (V, E, \textbf{X}_V) \), where \( V \in \mathbb{R}^{n} \) is a set of nodes , \( E \) is a set of undirected edges, and \( \textbf{X}_V \in \mathbb{R}^{n\times d}\) as the feature matrix. Each PFAS was represented as a node \( v \in V \) with a feature vector $\textbf{x}_{vi} \in \mathbb{R}^d$. The labeled nodes are denoted as \( L \) with labels, \( y_L \). The goal of the GCN is to predict the labels (binding affinity) of \( y_U \) for the unlabeled PFAS, \( U = V \setminus L \), which consist of the PFAS from the OECD dataset \cite{b22}.

\subsubsection{Graph representation}

In the proposed method, individual PFAS molecules were used as nodes. Each node contains a feature vector, \( \textbf{x}_{vi} \), which constitutes the calculated physical-chemical properties. To build edges between PFAS, we utilized the PFAS molecular structure information. The similar structured molecules have connections added to pass physical-chemical properties. We selected each node's $K$ nearest neighbor to add edges.  We used cosine similarity between PFAS to construct $K$ minimum edges per node. So as not to discriminate between equally similar PFAS, $K$ was set as a minimum number of edges to be formed. Thus, edges were formed for a node if it was within the top $K$ similarities. The cosine similarities were calculated using the molecular fingerprints between two nodes (PFAS) \( u \) and \( v \) defined as:

\begin{equation}
\text{cosine similarity(u,v)} = \frac{\sum_{i=1}^{n} u_i v_i}{\sqrt{\sum_{i=1}^{n} u_i^2} \sqrt{\sum_{i=1}^{n} v_i^2}}
\end{equation}

where, \( u_i \) and \( v_i \) are the calculated molecular fingerprints for the respective PFAS. In our experiments, we set $K=4$ to form edges. 

\subsubsection{Message Passing}

In our GCN architecture, we implement the message-passing operation using the GraphSAGE \cite{SAGE} framework. 
For each node \( v \in V \), the model aggregates information from its immediate neighbors \( N(v) \). The feature vectors of these neighboring nodes are aggregated using the mean function, as defined by:
\begin{equation}
\mathbf{h}_{N(v)}^{(k)} = \text{Mean}\left(\left\{\mathbf{h}_u^{(k-1)} \mid u \in N(v)\right\}\right)
\end{equation}
here, \( \mathbf{h}_u^{(k-1)} \) is the feature vector of neighbor \( u \) at layer \((k-1)\).

The aggregated neighborhood information \( \mathbf{h}_{N(v)}^{(k)} \) is combined with the node's own features \( \mathbf{h}_v^{(k-1)} \) through a linear transformation. This transformation is given by:

\begin{equation}
\mathbf{h}_v^{(k)} = \sigma\left(\text{BN}\left(\mathbf{W}_\text{self}\mathbf{h}_v^{(k-1)} + \mathbf{W}_\text{neigh}\mathbf{h}_{N(v)}^{(k)} + \mathbf{b}\right)\right)
\end{equation}

where \( \mathbf{W}_\text{self} \) and \( \mathbf{W}_\text{neigh} \) are the learnable weight matrices for the node's own features and the aggregated neighborhood features, respectively, and \( \mathbf{b} \) is the bias term. The batch normalization function \( \text{BN}(\cdot) \) \cite{batch_norm} is applied after the linear transformation and before the non-linear activation function \( \sigma(\cdot) \), which could be ReLU or Leaky ReLU depending on the molecular fingerprint.

\subsubsection{Feature Propagation and Layer Stacking}

The process of neighborhood aggregation, feature transformation, batch normalization, and activation is repeated across multiple layers of the GCN. To prevent overfitting and improve the generalization of the model, dropout was applied after the activation function in each layer. The final layer outputs binding affinity labels \( y_U \) for the unlabeled PFAS molecules. During the graph construction, disconnected sub-graphs were not removed from the overall network unless they only contained unlabeled nodes, as they would not learn from the training process. The model was trained using the mean squared error loss function (MSE) as shown below,
\begin{equation} \label{MSE}
\text{MSE} = \frac{1}{|L|} \sum_{v \in L} \left( \hat{y}_v - y_v \right)^2
\end{equation}
where ${|L|}$ is the number of labeled nodes, $\hat{y}_v$ is the predicted binding affinity for node $v$, and $y_v$ is the true binding affinity.

\subsection{Uncovering of Binding Mechanisms}

\subsubsection{Clustering}
To expand upon the ability to predict the unknown binding affinities of PFAS to LFABP, the PFAS in our dataset were then clustered by combining the predicting binding affinities along with the molecular fingerprint. Due to the semi-supervised nature of the model, the bulk of the valuable information lies in grouping PFAS to identify patterns and similarities among PFAS characteristics. Thus, clustering the PFAS provides a method to evaluate the unlabeled PFAS. After evaluating various clustering methods, the Ward algorithm was chosen for its best performance by finding the convergence of three different scoring functions: Silhouette score, Davies-Bouldin index, and the Calinski-Harabasz index.

\subsubsection{MD simulation}
To investigate the binding mechanisms of selected PFAS, we employed MD simulations. The specific PFAS compounds chosen for this study were from the final dataset and individually chosen from clusters from the above method. In alignment with the experimental data referenced \cite{b16}, the PFAS were placed in a simulation environment with LFABP, complexed with oleic acids (PDB code: 2LKK), downloaded from the RCSB website \cite{2lkk}. The PFAS were built using CHARMM-GUI's \cite{charmm1,charmm2} Ligand Modeler \cite{ligand}. The simulation was further processed by combining LFABP and the PFAS using Solution Builder \cite{sol_build}, and Multicomponent Assembler \cite{multicomp}, also tools from CHARMM-GUI. Three simulations were performed for each PFAS, with a PFAS randomly placed around a LFABP complex in a 60 x 60 x 60 angstrom (\r{A}) cubic box, maintaining at least 15 \r{A} of center of mass (COM) distance between the PFAS and the LFABP. The simulation solvent contained TIP3P water molecules with potassium and chloride ions (KCl) to neutralize the system.  All simulations were performed following system minimization and equilibrations (details provided in the following section). Before initiating the production run, we removed the oleic acids from the binding site within the protein to expose the non-portal and portal binding sites for the PFAS. This step was performed in accordance with the previous study \cite{b16} that provided the PFAS binding affinity training data. The system was re-neutralized before the production simulation.

\section{Experimental Setup}\label{sec:experiment}

\subsection{Data}
A dataset of PFAS and calculated docking scores from previous research \cite{b16} was adopted for this study to train the GCN model. The aforementioned group used molecular dynamics simulations to calculate docking scores of non-portal region of LFABP as binding affinity. We utilized $829$ data points generated by their study.
To generalize the study into broader PFAS groups, we combined unlabeled OECD dataset with labeled dataset to obtain binding affinity through our proposed semi-supervised GCN model. The OECD dataset contained $4730$ PFAS\cite{b22}. After filtering for PFAS with SMILES strings, we utilized $2284$ PFAS from the OECD database in our study.

To optimize the proposed GCN model, we split the labeled dataset into 60\% training, 20\% validation and 20\% testing. The same random seed was assigned to split the dataset for training the proposed model and baseline models. 

\subsection{Molecular Fingerprint and Descriptor Calculation}

Using the online-tool ChemDes \cite{b24}, $12$ molecular fingerprints were obtained for each unique PFAS. The molecular fingerprints consisted of the following: AtomPairs2D fingerprint (AP2D, 780 bits), AtomPairs2D count fingerprint (AP2D\_C, 780 bits), CDK fingerprint (CDK, 1024 bits), CDK extended fingerprint (CDKE, 1024 bits), Graphonly fingerprint (CDK\_g, 1024 bits), Klekota-Roth fingerprint (KR, 4860 bits), Klekota-Roth count fingerprint (KR\_C, 4860 bits), MACCS fingerprint (MACCS, 166 bits), PubChem fingerprint (PC, 881 bits), Substructure fingerprint (Sub, 307 bits), and Substructure count fingerprint (Sub\_C, 307 bits). In addition to the fingerprints, molecular descriptors were calculated using Mordred \cite{b25}. Initially, $1753$ categories of descriptors were calculated and reduced to $705$ by removing high inter-correlated features (pairwise correlation coefficient $>0.95$) due to the redundancy and lack of important information contained in these inter-correlated features. The resulting fingerprints and descriptors were standardized feature-wise with the following method:

\begin{equation}
    X_i = \frac{X_0 - \mu}{\sigma}
\end{equation}

where \( X_i \) is the standardized value of fingerprints or descriptors, \( X_0 \) is the original value, \( \mu \) is the feature mean, and \( \sigma \) is the feature standard deviation.

\subsection{Baseline Models}

To benchmark the best performing GCN models, we compared it against several baseline machine learning models. The features used for these models were derived from a combination of Mordred descriptors and a selected fingerprint from a pool of twelve, with each model testing a total of $12$ combinations of descriptor and fingerprint. The hyper-parameters for the following models were optimized with Scikit-learn’s grid search cross validation method with ten splits due to the limited dataset \cite{b26}. The same hyper-parameters used in a related study were applied here for consistency in comparison \cite{b16}. The coefficient of determination ($R^2$) was used to evaluate the performance of the results obtained from the machine learning algorithms.

\textit{RF. } The performance of the graph model was compared with that of the random forest (RF) algorithm. The following hyper-parameters were adjusted during the grid search process: max depth, max features, minimum samples split, and number of estimators.

\textit{DT. } The performance of the proposed graph model was compared with that of the decision tree (DT) algorithm. The following hyper-parameters were adjusted during the grid search process: max depth and minimum samples leaf.

\textit{Ridge. }  The performance of the proposed graph model was compared with that of the Ridge regression algorithm. Only the alpha hyper-parameter was optimized.

\textit{SVR. }  The performance of the proposed graph model was compared with that of the support vector regression (SVR) algorithm. The following hyper-parameters were adjusted during the grid search process: kernel and gamma.

\textit{DNN. } A shallow fully connected dense neural network (DNN) model was compared as well with the proposed graph model. The DNN was constructed with PyTorch’s layers \cite{b27} and utilized the Adam optimizer along with the MSE loss function (\ref{MSE}) . In order to serve a regression task, the last layer of every model only contained one node. The following hyper-parameters were adjusted during the grid search process: number of hidden layer nodes, number of hidden layers, weight normalization, learning rate, activation function, dropout, and batch size.

\subsection{MD setup}

All experiments were simulated with Gromacs $2024$ \cite{gromacs}.  The simulation temperature was maintained at $300$ K. All simulations were performed following system minimization and equilibrations. We performed $500,000$ steps of minimization using the steepest descent algorithm. We performed both NVT (constant volume, temperature, and number of particles) and NPT (constant pressure, temperature, and number of particles) equilibration for $1$ ns with $1$ fs of time step, using V-rescale \cite{rescale} for temperature coupling with $1$ ps of time constant and C-rescale \cite{rescale} for pressure coupling with $2$ ps of time constant. The final simulation was executed at $300$ K and a pH of $7$ for a duration of $1000$ ns with trajectory frames extracted at $250$ ps intervals. COM distances were calculated between the PFAS and the potential binding sites of LFABP, the nonportal and portal regions. Simulation visuals were created using Pymol \cite{pymol}.

\section{Results \& Discussion}\label{sec:result}

\subsection{Overall Results}

Initially in the study, the performance of the machine learning models were evaluated using one fingerprint (AtomPairs2D count) in combination with Mordred descriptors to predict the PFAS binding affinity towards LFABP. Table~\ref{tab:AP2D_model} shows the improvement in model performance when utilizing a more complex deep learning models as shown by the DNN ($R^2=0.62$) and GCN ($R^2=0.66$). Even though the DNN performs better than the other baseline models, the proposed GCN shows greater potential stemming from its ability to use both the fingerprint and descriptor while avoiding overfitting. This is due to the proposed model's ability to apply the same information as the DNN but reduce the number of node features. The GCN also benefits from constructing edges between only PFAS that are more similar to each other. This particular GCN model uses AP2D count to construct the graph while using the Mordred descriptors as the feature set.

\begin{table}[h]
\caption{AP2D Count and Mordred Model Performance}
    \centering
    \begin{tabular}{|p{1.8cm}|p{3.0cm}|} \hline
    Model& Model Performance ($R^2$)\\\hline\hline
    Ridge& 0.53\\ \hline
    DT& 0.54\\ \hline
    SVR& 0.58\\ \hline
    RF& 0.60\\ \hline
    DNN& 0.62\\ \hline
    GCN& 0.66\\\hline
    \end{tabular}
    \label{tab:AP2D_model}
\end{table}

\begin{figure}[htbp]
    \centering
    \includegraphics[width=0.9\columnwidth]{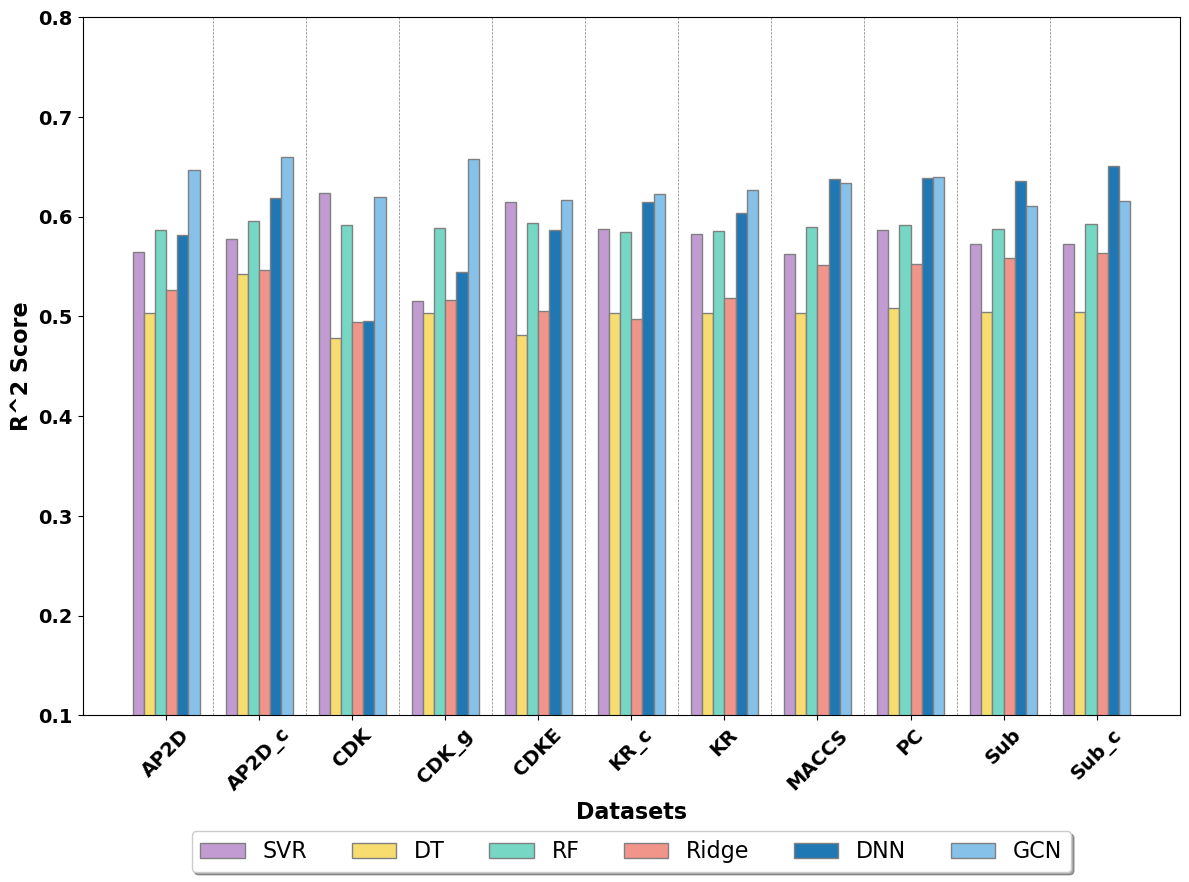}
    \caption{Performance of binding affinity prediction models of PFAS docking into LFABP. }
    \label{fig:final_r2s}
    \vspace{-0.1in}
\end{figure}

To confirm the use of AP2D count as the best performing fingerprint, potential combinations of fingerprints with the Mordred descriptors were employed for each model (shown in Figure \ref{fig:final_r2s}). A total of $66$ models were compared against each other as the best performing of its specific categorization %(Fig.~\ref{fig:final_r2s}). 
Every model used Mordred descriptors in addition to a single molecular fingerprint. While the GCN processed one of the feature sets for constructing a graph and the other as features, the other models combined both for the feature set. The GCN's showed greater capability to predict the binding affinity when utilizing the fingerprint to construct the graph and the descriptors as the feature set, as such this process is what is shown in Figure~\ref{fig:final_r2s} for the GCN's. The GCN performed better for the majority fingerprints while the DNN was able to compete for the other fingerprints. The best performing model was the proposed GCN using the AP2D count fingerprint to calculate similarity scores building a graph and the Mordred descriptors as the feature set. This combination achieved an external validation $R^2$ score $0.66$, outperforming the previously reported best performing model in \cite{b16} by $0.3$.

\subsection{Clustering results}

\begin{figure}[htbp]
    \centering
    \includegraphics[width=0.85\columnwidth]{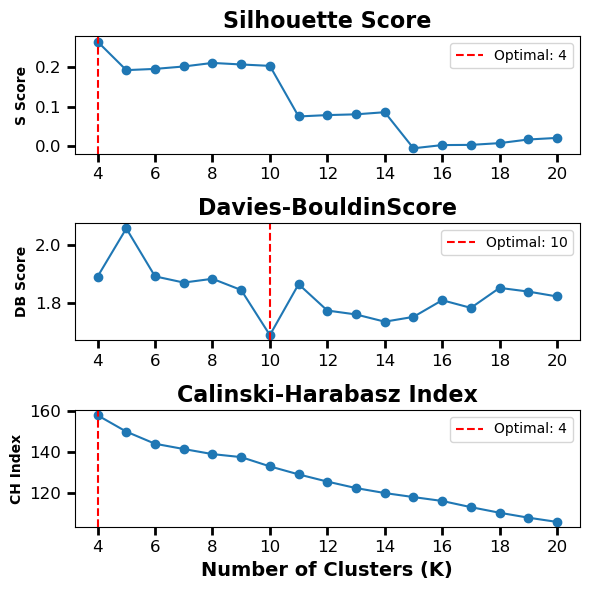}
    \caption{Unsupervised scoring function comparison to determine $K$ optimal clusters for the Ward algorithm.}
    \label{fig:ward_k_opt}
    \vspace{-0.1in}
\end{figure}

\begin{figure}[htbp]
    \centering
    \includegraphics[width=\columnwidth]{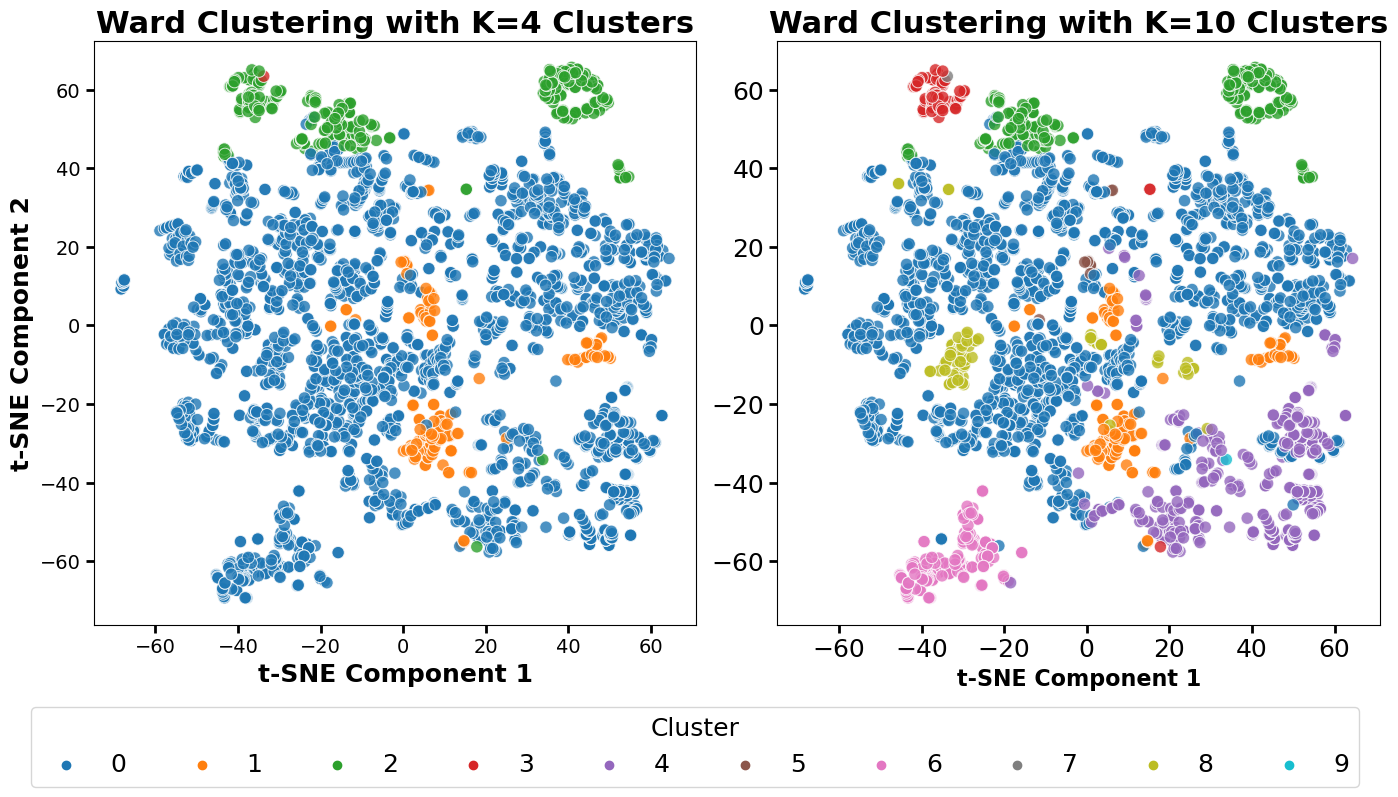}
    \caption{Global PFAS Ward cluster visualization utilizing t-SNE.}
    \label{fig:cluster_viz}
    \vspace{-0.1in}
\end{figure}

To expand upon the ability to predict the unknown binding affinities of PFAS to LFABP, the PFAS in our dataset were clustered by combining the predicting binding affinities along with the AP2D count fingerprint, as utilized by the proposed GCN model . After evaluating various clustering methods, the Ward algorithm was chosen for its performance using three different scoring functions; Silhouette (S) score, Davies-Bouldin (DB) index, and the Calinski-Harabasz (CH) index. These scoring methods also determined the optimal number of clusters, $K$. Figure~\ref{fig:ward_k_opt} shows the convergence at  $K=10$, indicated by the lowest valley for the DB score and peak at $10$ for the S score. Further confirmation of the optimal number of clusters is provided by visualizing the created clusters for $K=4$ and $K=10$, as shown in Figure \ref{fig:cluster_viz}.  The $K=10$ clustering reveals distinct clusters that are not apparent in the $K=4$ clustering. 

The clusters were further analyzed to understand the distribution of docking scores and other key characteristics of PFAS within each cluster. The characteristics chosen for analysis included the longest fluorinated carbon chain length, the number of fluorinated carbons, and the presence of functional groups, as these factors are critical for distinguishing PFAS groups. 

\begin{figure}[htbp]
    \centering
    \includegraphics[width=\columnwidth]{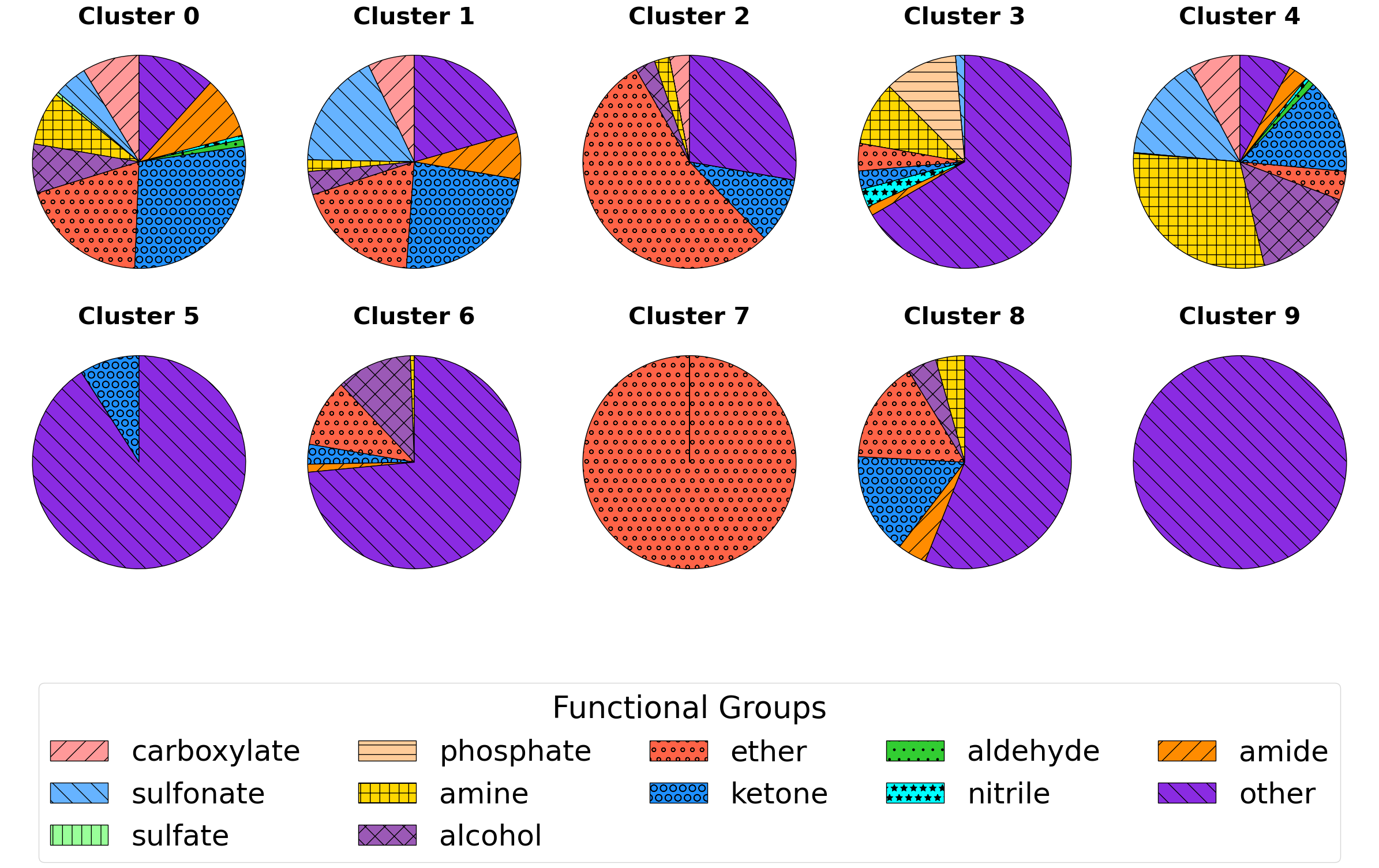}
    \caption{Presence of functional groups in PFAS distributed among clusters.}
    \label{fig:funct_groups}
\end{figure}

\begin{figure}[htbp]
    \centering
    \includegraphics[width=0.8\columnwidth]{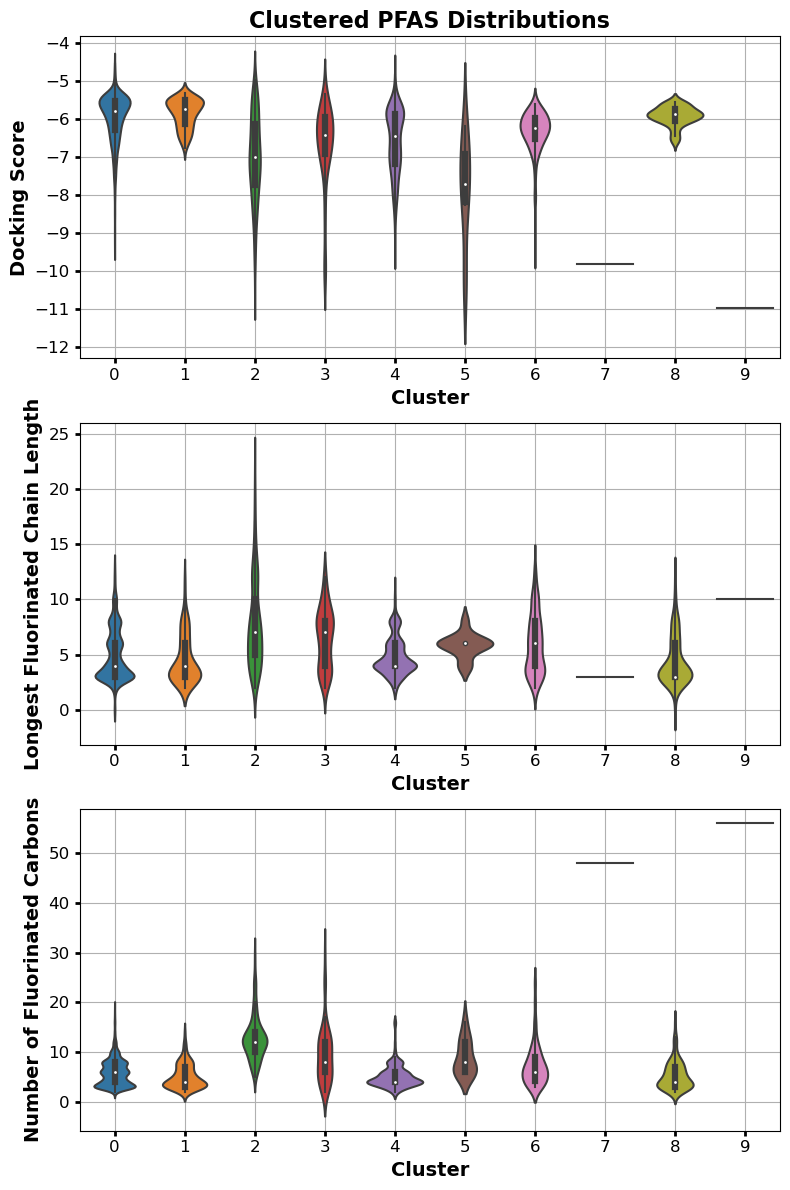}
    \caption{Distributions of the clustered PFAS based on the Ward algorithm. The best performing GCN predicted the docking scores of the global dataset of PFAS.}
    \label{fig:cluster_dist}
    \vspace{-0.1in}
\end{figure}

Notably, clusters $7$ and $9$ were formed each with a single PFAS. This distinction is of note as it picks out two of the most extreme PFAS in the dataset, both shown in Figure~\ref{fig:funct_groups}. The key marker for these PFAS is the large number of fluorinated carbons present and the fact that they are not represented in a single long chain, but multiple medium length chains as shown in Figure~\ref{fig:cluster_dist}. Such is one reason that this parameter was given distinction from the longest CF chain length parameter, Figure~\ref{fig:cluster_dist}. Another example for this necessary distinction would be the newer GenX PFAS which breaks up the standard long continuous CF chain into two different components, thus having a similar number of fluorinated carbons to legacy long chain PFAS  while reducing the longer fluorinated carbon chains.

\subsection{MD simulation}

\begin{figure}[htbp]
    \centering
    \includegraphics[width=\columnwidth]{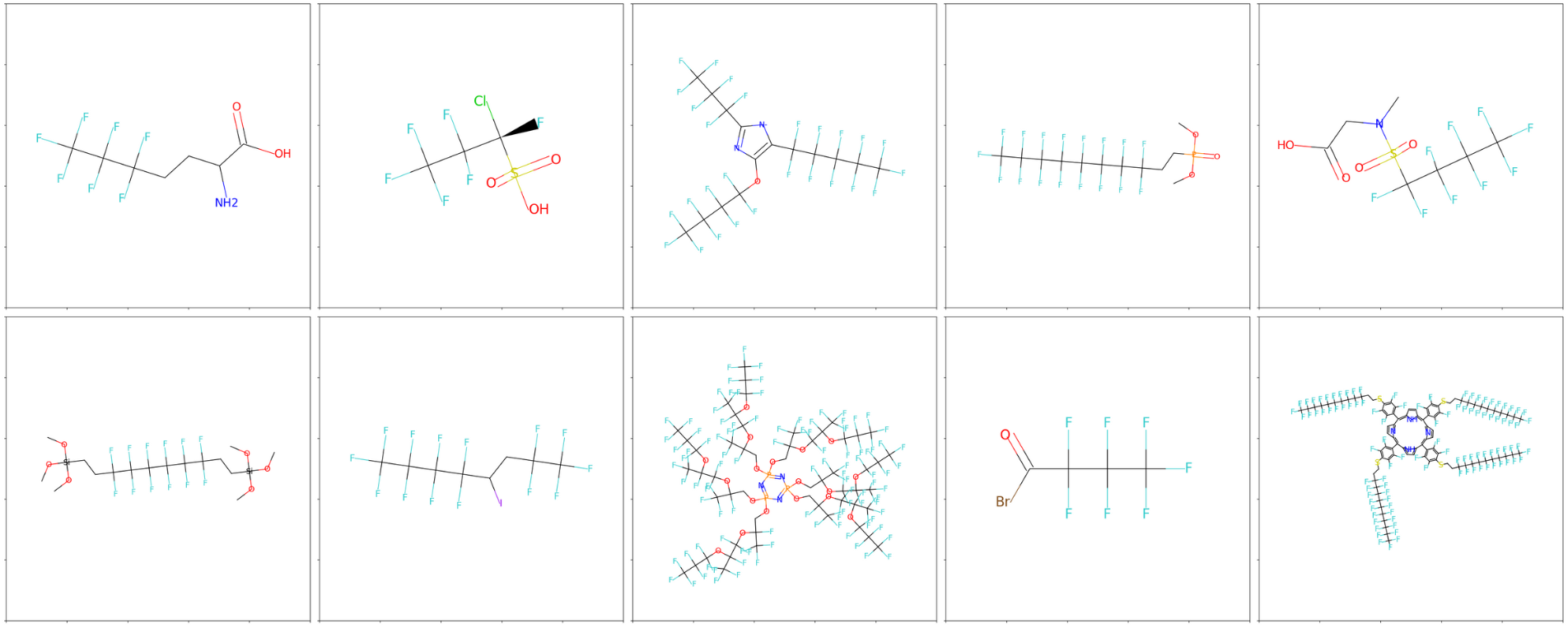}
    \caption{Representative PFAS from each cluster from the Ward algorithm. The PFAS belong to Clusters 0 through 9 starting from left to right, top to bottom.}
    \label{fig:clust_pfas}
    \vspace{-0.1in}
\end{figure}

The PFAS within each cluster were examined to identify representative PFAS that could be further isolated to reveal binding mechanisms to LFABP using MD simulations (see Figure~\ref{fig:clust_pfas}). A quick visual inspection of the PFAS chosen from each cluster shows the wide variations of PFAS that exist in industry. Due to the CF chain commonality, the clusters pick up on the functional groups and specific atoms as distinguishing elements to separate them from each other. These functional groups and atoms are of great significance when breaking down the specific binding interactions for PFAS. This because CF bonds typically having a strong affinity towards each other, such that they are not encouraged to create covalent bonds. 

\begin{figure}[htbp]
    \centering
    \includegraphics[width=\columnwidth]{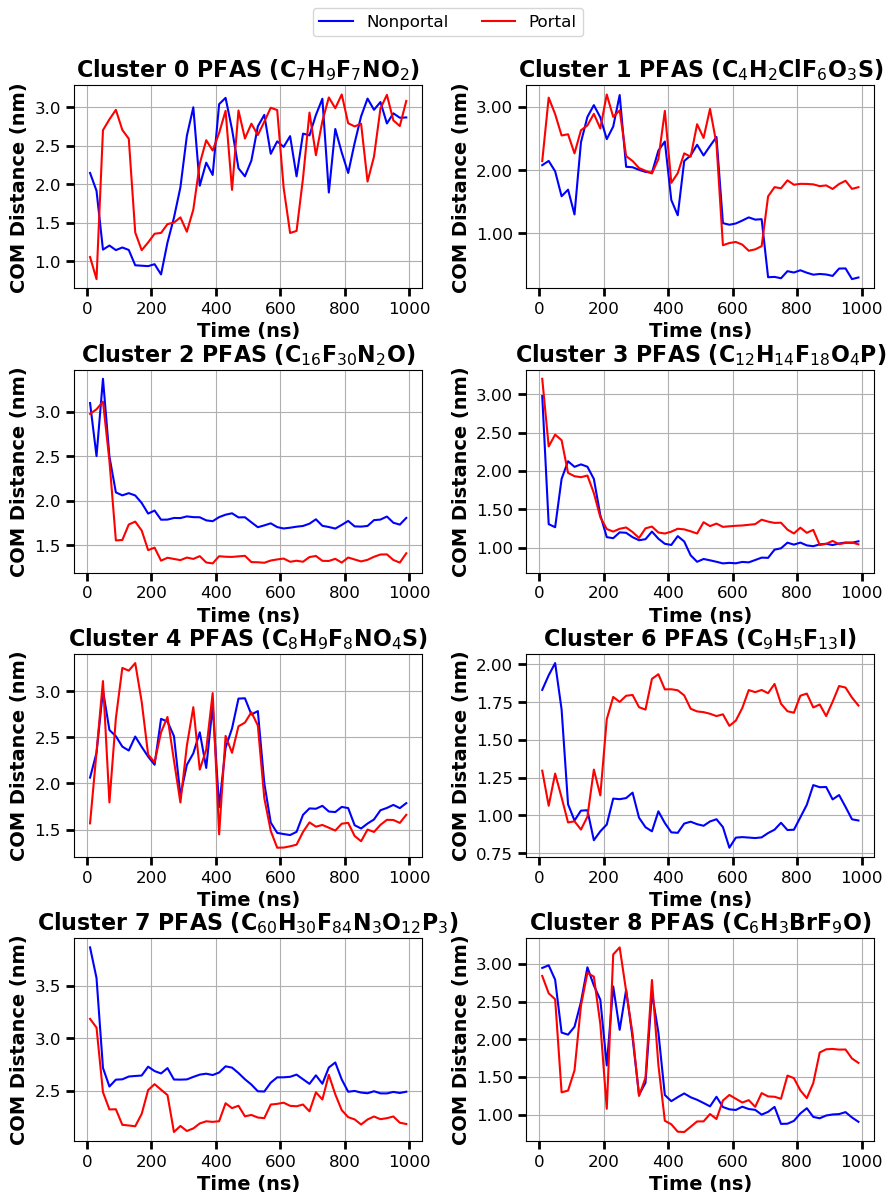}
    \caption{Center of mass distances between LFABP regions (nonportal and portal) and PFAS.}
    \label{fig:com_dist}
    
\end{figure}

Figure~\ref{fig:com_dist} shows the change in COM distances between the PFAS and the two binding regions of LFABP. The nonportal region consists of the following residues: ARG122 (arginine), SER39 (serine), and SER124. While the portal region contains these residues: LYS31 (lysine) and SER56. The nonportal residues lie within LFABP, while the portal residues are closer to the corner and outside of LFABP. The COM distances shown are one of the three repeated simulations conducted.
During the simulations, not all PFAS converged to a stable location when interacting with LFABP, specifically PFAS $0$ and $1$. 
This could be determined when a COM distances never stabilize such as the representative PFAS from cluster $0$ in Figure~\ref{fig:com_dist}. A trend starts to appear when analyzing the PFAS similar to that cluster, in terms of smaller molecules with short CF chains. Smaller molecules take longer to interact with the protein due to their smaller surface. The PFAS from clusters $0$, $1$, $4$, and $8$ take longer to interact due to this reason. Another contributing factor for this trend, is the driving force for initial contact with the LFABP appears to be the hydrophobic nature of the CF chains. This may push the PFAS to come into contact with the protein in order for functional groups to bond to the binding sites. \begin{figure}[htbp]
\vspace{-0.7in}
    \centering
    \begin{subfigure}[b]{0.35\textwidth}
        \centering
        \includegraphics[width=\textwidth]{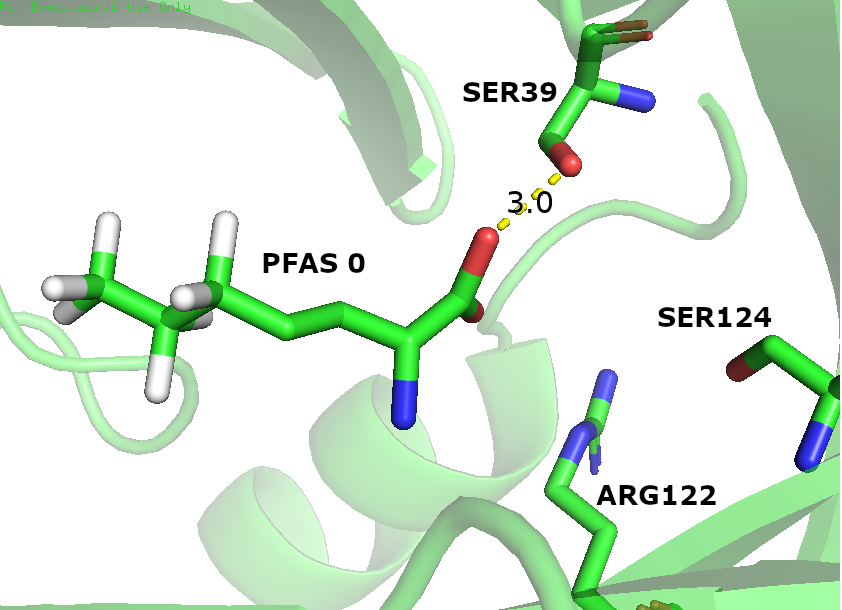}
        
    \end{subfigure}
    \hfill
    \begin{subfigure}[b]{0.35\textwidth}
        \centering
        \includegraphics[width=\textwidth]{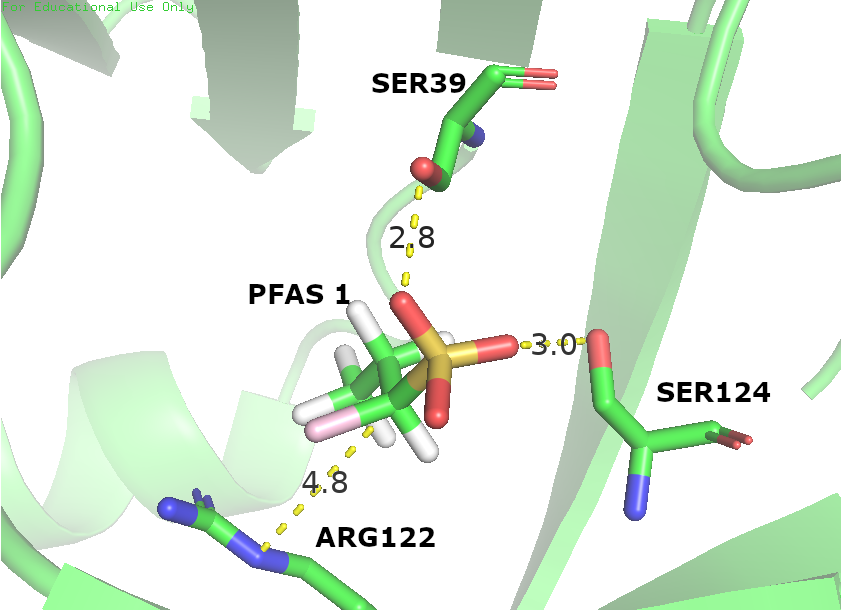}
        
    \end{subfigure}
    \vfill
    \begin{subfigure}[b]{0.35\textwidth}
        \centering
        \includegraphics[width=\textwidth]{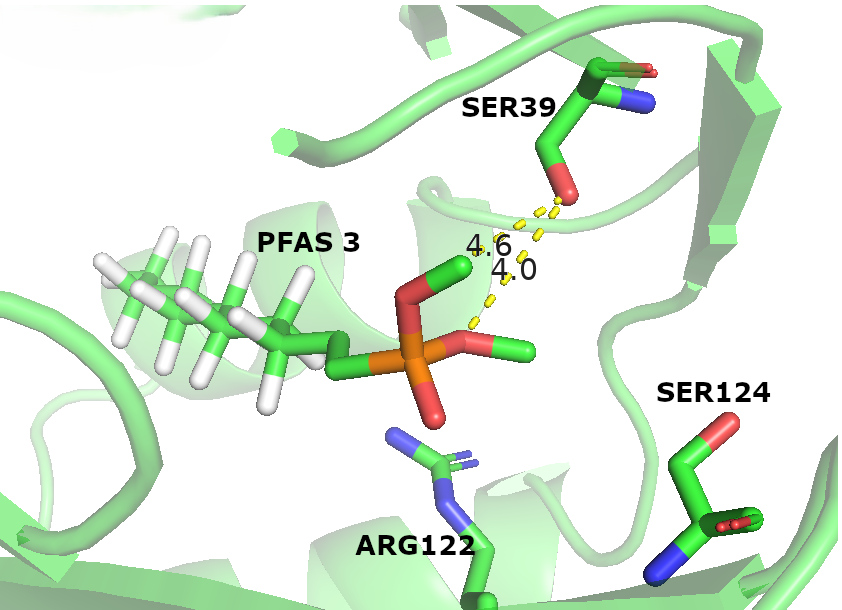}
        
    \end{subfigure}
    \hfill
    \begin{subfigure}[b]{0.35\textwidth}
        \centering
        \includegraphics[width=\textwidth]{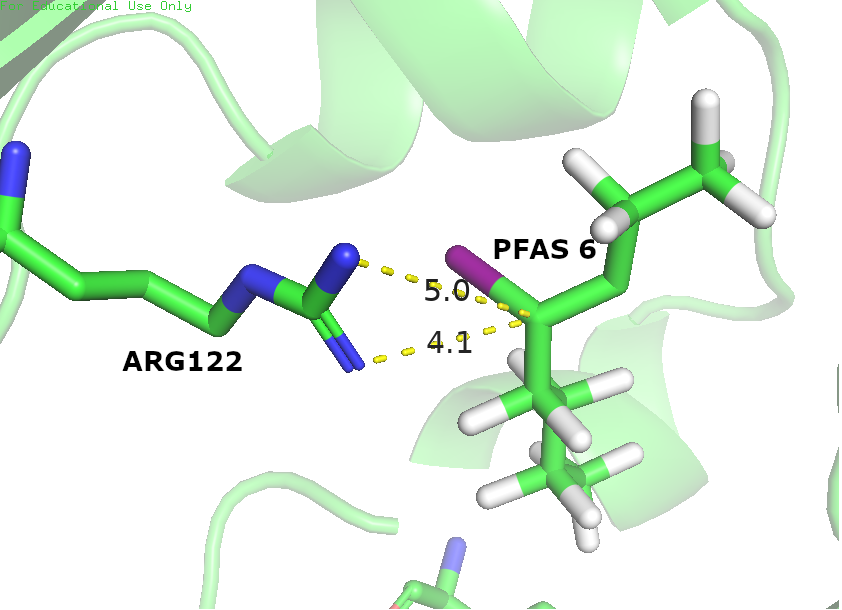}
        
    \end{subfigure}
    \caption{Potential binding interactions between PFAS and nonportal residues. Interactions are labelled with a yellow dotted lines, distances shown in \r{A}. Atoms are denoted by the following colors: carbon, green; oxygen, red; nitrogen, blue; phosphorus, orange; sulfur, yellow; bromine, purple; chlorine, pink; fluorine, white.}
    \label{fig:binding}
    \vspace{-0.1in}
\end{figure}This might also guide PFAS inside LFABP as less water is present on the inside of the protein. The larger chain lengths and more CF bonds cause PFAS to interact faster as can be seen specifically with the PFAS from clusters $2$, $3$, and $7$. For two of these PFAS (cluster $2$ and $7$) they do not actually bond to the residues as the functional groups are blocked by the longer and numerous CF chains. As opposed to the PFAS from cluster $3$, where it contains the traditional PFAS structure of a singular CF chain with a functional group head that has easy access for binding interactions.

\subsection{Mechanism Discussion}

Figure~\ref{fig:binding} shows the binding interactions of PFAS from clusters $0$, $1$, $3$, and $6$ with the residues in the nonportal region. These PFAS demonstrated potential binding interactions due to the proximity of atoms of interest between the PFAS and the residues. A potential binding site was determined if the atoms had the potential to form covalent bonds and if the distance between said atoms was less than 5\r{A} \cite{b16}. Among these PFAS, interactions were favored with the lone oxygen on SER39, especially interacting with oxygens on the PFAS, if present. The only PFAS of this group without an oxygen, PFAS 6, potentially bonded with the nitrogens on ARG122 through its carbon of interest. This mode of binding also appears in the simulation with PFAS 1, as its carbon (that is bonded to a chlorine) is within the cutoff distance of a nitrogen on ARG122. This suggests favorable interactions between carbons that are bonded to a single halogen and nitrogens.

\section{Conclusion}\label{sec:conclusion}

In this research, we have addressed the challenge of predicting PFAS hepatotoxicity with limited data by proposing a novel method using molecular fingerprints and descriptors. Traditional approaches in this domain often are limited by a lack of binding affinity data and the use of an individual molecular fingerprint or descriptor. Our proposed method leverages the capabilities of GCN models to connect similar PFAS based on their structural similarities. This allows the model to utilize both fingerprints and descriptors, enhancing the predictive power of the proposed model while reducing overfitting. 
Our results demonstrate that the proposed method outperforms baseline models across 11 potential fingerprint and descriptor combinations. To uncover the binding mechanisms of these PFAS, they were clustered based on their molecular fingerprint and representative PFAS were selected for molecular dynamics simulations. These simulations revealed that smaller PFAS, especially in terms of carbon-fluorine chain length and the number of carbon-fluorine bonds, are more inconsistent in whether they will interact with LFABP. In contrast, larger PFAS consistently come into contact with LFABP and thus have a higher chance of forming bonds with the protein. Future work will focus on testing our method in predicting PFAS toxicity to other susceptible proteins in the human body.

\vspace{12pt}


\begin{thebibliography}{00}
\bibitem{b1} Y. Liu, L. Wang, X. Zhang, Y. Zhang, L. Zhang, and X. Zheng, "A review of single-molecule sequencing technologies," *Journal of Biomedical Science*, vol. 26, no. 1, pp. 1-11, 2019. [Online]. Available: https://www.ncbi.nlm.nih.gov/pmc/articles/PMC6380916/. [Accessed: Aug. 8, 2024].

\bibitem{b2} J. Glüge et al., “An overview of the uses of per- and polyfluoroalkyl substances (PFAS),” Environ. Sci.: Processes Impacts, vol. 22, no. 12, pp. 2345–2373, 2020, doi: 10.1039/D0EM00291G.
\bibitem{b3} I. T. Cousins et al., “The high persistence of PFAS is sufficient for their management as a chemical class,” Environ. Sci.: Processes Impacts, vol. 22, no. 12, pp. 2307–2312, 2020, doi: 10.1039/D0EM00355G.
\bibitem{b4} C. Lau, K. Anitole, C. Hodes, D. Lai, A. Pfahles-Hutchens, and J. Seed, “Perfluoroalkyl Acids: A Review of Monitoring and Toxicological Findings,” Toxicological Sciences, vol. 99, no. 2, pp. 366–394, Oct. 2007, doi: 10.1093/toxsci/kfm128.
\bibitem{b5} A. M. Calafat, L.-Y. Wong, Z. Kuklenyik, J. A. Reidy, and L. L. Needham, “Polyfluoroalkyl Chemicals in the U.S. Population: Data from the National Health and Nutrition Examination Survey (NHANES) 2003–2004 and Comparisons with NHANES 1999–2000,” Environ Health Perspect, vol. 115, no. 11, pp. 1596–1602, Nov. 2007, doi: 10.1289/ehp.10598.
\bibitem{b6} L. W. Y. Yeung, C. Dassuncao, S. Mabury, E. M. Sunderland, X. Zhang, and R. Lohmann, “Vertical Profiles, Sources, and Transport of PFASs in the Arctic Ocean,” Environ. Sci. Technol., vol. 51, no. 12, pp. 6735–6744, Jun. 2017, doi: 10.1021/acs.est.7b00788.
\bibitem{b7} J. Dai, M. Li, Y. Jin, N. Saito, M. Xu, and F. Wei, “Perfluorooctanesulfonate and Perfluorooctanoate in Red Panda and Giant Panda from China,” Environ. Sci. Technol., vol. 40, no. 18, pp. 5647–5652, Sep. 2006, doi: 10.1021/es0609710.


\bibitem{b12} S. E. Fenton et al., “Per‐ and Polyfluoroalkyl Substance Toxicity and Human Health Review: Current State of Knowledge and Strategies for Informing Future Research,” Enviro Toxic and Chemistry, vol. 40, no. 3, pp. 606–630, Mar. 2021, doi: 10.1002/etc.4890.

\bibitem{b13} H. Cao et al., “Screening of Potential PFOS Alternatives To Decrease Liver Bioaccumulation: Experimental and Computational Approaches,” Environ. Sci. Technol., vol. 53, no. 5, pp. 2811–2819, Mar. 2019, doi: 10.1021/acs.est.8b05564.

\bibitem{b14} L. Zhang, X.-M. Ren, and L.-H. Guo, “Structure-Based Investigation on the Interaction of Perfluorinated Compounds with Human Liver Fatty Acid Binding Protein,” Environ. Sci. Technol., vol. 47, no. 19, pp. 11293–11301, Oct. 2013, doi: 10.1021/es4026722.

\bibitem{b15} W. Cheng and C. A. Ng, “Predicting Relative Protein Affinity of Novel Per- and Polyfluoroalkyl Substances (PFASs) by An Efficient Molecular Dynamics Approach,” Environ. Sci. Technol., vol. 52, no. 14, pp. 7972–7980, Jul. 2018, doi: 10.1021/acs.est.8b01268.

\bibitem{b16} J. Zhao et al., “Hepatotoxicity assessment investigations on PFASs targeting L-FABP using binding affinity data and machine learning-based QSAR model,” Ecotoxicology and Environmental Safety, vol. 262, p. 115310, Sep. 2023, doi: 10.1016/j.ecoenv.2023.115310.

\bibitem{b17} W. Cheng, J. A. Doering, C. LaLone, and C. Ng, “Integrative Computational Approaches to Inform Relative Bioaccumulation Potential of Per- and Polyfluoroalkyl Substances Across Species,” Toxicological Sciences, vol. 180, no. 2, pp. 212–223, Apr. 2021, doi: 10.1093/toxsci/kfab004.

\bibitem{b18} W. Cheng and C. A. Ng, “Using Machine Learning to Classify Bioactivity for 3486 Per- and Polyfluoroalkyl Substances (PFASs) from the OECD List,” Environ. Sci. Technol., vol. 53, no. 23, pp. 13970–13980, Dec. 2019, doi: 10.1021/acs.est.9b04833.

\bibitem{b19} T. T. Lai, D. Kuntz, and A. K. Wilson, “Molecular Screening and Toxicity Estimation of 260,000 Perfluoroalkyl and Polyfluoroalkyl Substances (PFASs) through Machine Learning,” J. Chem. Inf. Model., vol. 62, no. 19, pp. 4569–4578, Oct. 2022, doi: 10.1021/acs.jcim.2c00374.

\bibitem{b20} H. Cao et al., “Investigation of the Binding Fraction of PFAS in Human Plasma and Underlying Mechanisms Based on Machine Learning and Molecular Dynamics Simulation,” Environ. Sci. Technol., vol. 57, no. 46, pp. 17762–17773, Nov. 2023, doi: 10.1021/acs.est.2c04400.

\bibitem{b21} T. N. Kipf and M. Welling, “Semi-Supervised Classification with Graph Convolutional Networks,” arXiv:1609.02907 [cs, stat], Feb. 2017, arXiv: 1609.02907. [Online]. Available: http://arxiv.org/abs/1609.02907

\bibitem{b22} OECD (Organisation for Economic Co-operation and Development), "Summary report on the new comprehensive global database of Per- and Polyfluoroalkyl Substances (PFASs)," Publications Series on Risk Management No. 39, 2018. [Online]. Available: \url{http://www.oecd.org/officialdocuments/publicdisplaydocumentpdf/?cote=ENV-JM-MONO(2018)7&doclanguage=en}

\bibitem{SAGE} W. L. Hamilton, R. Ying, and J. Leskovec, “Inductive Representation Learning on Large Graphs,” 2017, arXiv. doi: 10.48550/ARXIV.1706.02216.

\bibitem{batch_norm} S. Ioffe and C. Szegedy, “Batch Normalization: Accelerating Deep Network Training by Reducing Internal Covariate Shift,” 2015, arXiv. doi: 10.48550/ARXIV.1502.03167.

\bibitem{2lkk} J. Cai, C. Lücke, Z. Chen, Y. Qiao, E. Klimtchuk, and J. A. Hamilton, “Solution Structure and Backbone Dynamics of Human Liver Fatty Acid Binding Protein: Fatty Acid Binding Revisited,” \textit{Biophysical Journal}, vol. 102, no. 11, pp. 2585–2594, Jun. 2012, doi: \href{https://doi.org/10.1016/j.bpj.2012.04.039}{10.1016/j.bpj.2012.04.039}. 

\bibitem{charmm1} S. Jo, T. Kim, V. G. Iyer, and W. Im, “CHARMM-GUI: A web-based graphical user interface for CHARMM,” *J. Comput. Chem.*, vol. 29, pp. 1859–1865, 2008.

\bibitem{charmm2} B. R. Brooks et al., “CHARMM: The biomolecular simulation program,” *J. Comput. Chem.*, vol. 30, pp. 1545–1614, 2009.

\bibitem{ligand} S. Kim et al., “CHARMM-GUI ligand reader and modeler for CHARMM force field generation of small molecules,” *J. Comput. Chem.*, vol. 38, pp. 1879–1886, 2017.

\bibitem{sol_build} J. Lee et al., “CHARMM-GUI input generator for NAMD, GROMACS, AMBER, OpenMM, and CHARMM/OpenMM simulations using the CHARMM36 additive force field,” *J. Chem. Theory Comput.*, vol. 12, pp. 405–413, 2016.

\bibitem{multicomp} N. R. Kern et al., “CHARMM-GUI multicomponent assembler for modeling and simulation of complex multicomponent systems,” *bioRxiv*, 2023, doi: 10.1101/2023.08.30.555590.

\bibitem{b24} J. Dong et al., “ChemDes: an integrated web-based platform for molecular descriptor and fingerprint computation,” J Cheminform, vol. 7, no. 1, p. 60, Dec. 2015, doi: 10.1186/s13321-015-0109-z.

\bibitem{b25} H. Moriwaki, Y.-S. Tian, N. Kawashita, and T. Takagi, “Mordred: a molecular descriptor calculator,” J Cheminform, vol. 10, no. 1, p. 4, Dec. 2018, doi: 10.1186/s13321-018-0258-y.

\bibitem{b26} F. Pedregosa et al., "Scikit-learn: machine learning in python," *Journal of Machine Learning Research*, vol. 12, pp. 2825-2830, 2011.

\bibitem{b27} A. Paszke *et al.*, “PyTorch: An imperative style, high-performance deep learning library,” *arXiv preprint arXiv:1912.01703*, Dec. 2019. [Online]. Available: https://arxiv.org/abs/1912.01703. [Accessed: Aug. 8, 2024].

\bibitem{gromacs} M. J. Abraham et al., “GROMACS: High performance molecular simulations through multi-level parallelism from laptops to supercomputers,” *SoftwareX*, vol. 1–2, pp. 19–25, 2015, ISSN 2352-7110, doi: 10.1016/j.softx.2015.06.001.

\bibitem{rescale} G. Bussi, D. Donadio, and M. Parrinello, “Canonical sampling through velocity rescaling,” The Journal of Chemical Physics, vol. 126, no. 1, p. 014101, Jan. 2007, doi: 10.1063/1.2408420.

\bibitem{pymol} Schrödinger, LLC, “The PyMOL Molecular Graphics System, Version 1.8,” Nov. 2015. 

\bibitem{b29} MarvinSketch (version 24.1.2 , calculation module developed by ChemAxon, http://www.chemaxon.com/products/marvin/marvinsketch/, 2024.

\bibitem{35} O'Boyle et al. Open Babel: An open chemical toolbox. J Cheminform 3, 33 (2011). https://doi.org/10.1186/1758-2946-3-33

\bibitem{b23} National Cancer Institute, "Chemical Structure Lookup," Accessed: Aug. 07, 2024. [Online]. Available: https://cactus.nci.nih.gov/chemical/structure

\end{thebibliography}
\end{document}